\def\year{2021}\relax
\documentclass[letterpaper]{article} 
\usepackage{aaai21_2}  
\usepackage[hyphens]{url}  
\usepackage{graphicx} 
\usepackage{natbib}  
\usepackage{booktabs}       
\usepackage{microtype}      
\usepackage{algorithm}
\usepackage{algorithmic}
\usepackage{relsize}
\usepackage{array}
\usepackage{adjustbox}
\newcolumntype{d}[1]{D{.}{.}{#1}}
\newcolumntype{P}[1]{>{\centering\arraybackslash}p{#1}}
\usepackage{times}  
\usepackage{helvet} 
\usepackage{courier}  
\usepackage{caption} 
\usepackage{amsfonts}       
\usepackage{nicefrac}       
\usepackage{epsfig}
\usepackage{amsmath}
\usepackage{amssymb}
\usepackage{multirow}

\DeclareMathOperator*{\argmax}{arg\,max}
\DeclareMathOperator*{\argmin}{arg\,min}

\newtheorem{defn}{Definition}

\newtheorem{Lem}{Lemma}

\newtheorem{Rem}{Remark}

\newcommand{\oo}{{\mathbf o}}
\newcommand{\RR}{{\mathbb R}}
\newcommand{\xx}{{\mathbf x}}
\newcommand{\yy}{{\mathbf y}}

\newcommand{\Sbold}{{\mathbf S}}
\newcommand{\sbold}{{\mathbf s}}

\newcommand{\II}{{\mathbf I}}
\newcommand{\one}{{\mathbf 1}}

\newcommand{\XX}{{\mathbf X}}

\newcommand{\abold}{{\mathbf a}}
\newcommand{\bbold}{{\mathbf b}}
\newtheorem{Defn}{Definition}
\newtheorem{Prop}{Proposition}

\title{Variational Fair Clustering }
\author{
    Imtiaz Masud Ziko\textsuperscript{\rm 1},
    Jing Yuan\textsuperscript{\rm 2},
    Eric Granger\textsuperscript{\rm 1} and
    Ismail Ben Ayed\textsuperscript{\rm 1}
    \\
}
\affiliations{
    \textsuperscript{\rm 1} \'ETS Montreal, Canada \\
    \textsuperscript{\rm 2} Xidian University, China \\
}

\begin{document}
\maketitle

\begin{abstract}


We propose a general variational framework of fair clustering, which integrates an original Kullback-Leibler (KL) fairness term with a large class of clustering objectives, including prototype or graph based. Fundamentally different from the existing combinatorial and spectral solutions, our variational multi-term approach enables to control the trade-off levels between the fairness and clustering objectives. We derive a general tight upper bound based on a concave-convex decomposition of our fairness term, its Lipschitz-gradient property and the Pinsker's inequality. Our tight upper bound can be jointly optimized with various clustering objectives, while yielding a scalable solution, with convergence guarantee. Interestingly, at each iteration, it performs an independent update for each assignment variable. Therefore, it can be easily distributed for large-scale datasets. This scalability is important as it enables to explore different trade-off levels between the fairness and clustering objectives. Unlike spectral relaxation, our formulation does not require computing its eigenvalue decomposition. We report comprehensive evaluations and comparisons with state-of-the-art methods over various fair-clustering benchmarks, which show that our variational formulation can yield highly competitive solutions in terms of fairness and clustering objectives\footnote{Code is available at: \url{https://github.com/imtiazziko/Variational-Fair-Clustering}}.  
\end{abstract}

\section{Introduction}
\label{sec:introduction}
Machine learning models are impacting our daily life, for instance, in marketing, finance, education, and even in sentencing recommendations \cite{kleinberg2017human}. However, these models may exhibit biases towards specific demographic groups due to, for instance, the biases that exist within the data. For example, a higher level of face recognition accuracy may be found with white males \cite{buolamwini2018gender}, and a high probability of recidivism tends to be incorrectly predicted for low-risk African-Americans~\cite{Julia:2016}. These biases have recently triggered substantial interest in designing fair algorithms for the supervised learning  setting \cite{Hardt2016,Zafar2017,Donini2018}. Also, very recently, the community started to investigate fairness constraints in unsupervised learning \cite{Chierichetti2017,Kleindessner2019,backurs2019scalable,Samadi2018,Celis2018}. Specifically, Chierichetti et al. \cite{Chierichetti2017} pioneered the concept of {\em fair clustering}. The problem consists of embedding fairness constraints that encourage clusters to have balanced demographic groups pertaining to some sensitive attributes (e.g., sex, gender, race, etc.), so as to counteract any form of data-inherent bias. 

Assume that we are given $N$ data points to be assigned to a set of $K$ clusters, and let $S_k \in \{0,1\}^N$ denotes a binary indicator vector whose components take value $1$ when the point is within cluster $k$, and $0$ otherwise. Also suppose that the data contains $J$ different demographic groups, with $V_j \in\{0,1\}^N$ denoting a binary indicator vector of  demographic group $j$. The authors of
\cite{Chierichetti2017,Kleindessner2019} suggested to evaluate fairness in terms of cluster-balance measures, which take the following form:
\begin{eqnarray}
\label{cluster-balance}
\mbox{balance}(S_k) = \min_{j\neq j^{'}} \frac{V_j^tS_k}{V_{j^{'}}^tS_k}\in[0,1]
\label{eq:balance}
\end{eqnarray}
The higher this measure, the fairer the cluster. The overall clustering balance is defined by the minimum of Eq. \eqref{cluster-balance} over $k$. This notion of fairness in clusters has recently given rise to a new line of research that was introduced, mostly, for prototype-based clustering, e.g., K-center, K-median and K-means \cite{Chierichetti2017, backurs2019scalable,Schmidt2018, bera2019fair}. Also, very recently, fairness has been investigated in the context of spectral graph clustering \cite{Kleindessner2019}. The general problem raises several interesting questions. How to embed fairness in popular clustering objectives? Can we control the trade-off between some ``acceptable” fairness level (or tolerance) and the quality of the clustering objective? What is the cost of fairness with respect to the clustering objective and computational complexity? 

Chierichetti et al. \cite{Chierichetti2017} investigated combinatorial approximation algorithms
for maximizing the the fairness measures in Eq. \eqref{cluster-balance}, for K-center and K-median clustering, and for binary demographic groups ($J =2$). They compute {\em fairlets}, which are groups of points that are fair, and can not be split further into more subsets that are also fair. Then, they consider each fairlet as a data point, and cluster them with approximate K-center or K-median algorithms. Unfortunately, as reported in the experiments in \cite{Chierichetti2017}, obtaining fair solutions with these fairlets-based algorithms comes at the price of a substantial increase in the clustering objectives. Also, the cost for finding fairlets with perfect matching is quadratic w.r.t the number of data points, a complexity that increases for more than two demographic groups. Several combinatorial solutions followed-up on the work in \cite{Chierichetti2017} to reduce this complexity. For instance, Backurs et al. \cite{backurs2019scalable} proposed a solution to make the fairlet decomposition in \cite{Chierichetti2017} scalable for $J=2$, by embedding the input points in a tree metric. Rösner and Schmidt \cite{Rosner2018PrivacyPC} designed a 14-approximate algorithm for fair K-center. \cite{Schmidt2018,huang2019coresets} proposed fair K-means/K-median based on coreset -- a reduced proxy set for the full dataset. Bera et al. \cite{bera2019fair} provided a bi-criteria approximation algorithm for fair prototype-based clustering, enabling multiple groups ($J>2$). It is worth noting that, for large-scale data sets, \cite{Chierichetti2017,Rosner2018PrivacyPC,bera2019fair} sub-sample the inputs to mitigate the quadratic complexity w.r.t $N$. More importantly, the combinatorial algorithms discussed above are tailored for specific prototype-based objectives. For instance, they are not applicable to the very popular graph-clustering objectives, e.g., Ratio Cut or Normalized Cut \cite{VonLuxburg2007}, which limits applicability in a breadth of graph problems, in which data takes the form of pairwise affinities.

Kleindessner et al. \cite{Kleindessner2019} integrated fairness into graph-clustering objectives. They embedded linear constraints on the assignment matrix in spectral relaxation. Then, they solved a constrained trace optimization via finding the $K$ smallest eigenvalues of some transformed Laplacian matrix. However, it is well-known that spectral relaxation has heavy time and memory loads since it
requires storing an $N \times N$ affinity matrix and computing its eigenvalue decomposition – the complexity is cubic w.r.t $N$ for a straightforward implementation, and super-quadratic for fast implementations \cite{Tian-AAAI}. In the general context of spectral relaxation and graph partitioning, issues related to computational scalability for large-scale problems is driving an active line of recent work \cite{shaham2018spectralnet, ziko2018scalable,Vladymyrov-2016}.

The existing fair clustering algorithms, such as the combinatorial or spectral solutions discussed above, do not have mechanisms that control the trade-off levels
between the fairness and clustering objectives. Also, they are tailored either to prototype-based \cite{backurs2019scalable,bera2019fair,Chierichetti2017,Schmidt2018} or graph-based objectives \cite{Kleindessner2019}. Finally, for a breadth of problems of wide interest, such as pairwise graph data, the computation and memory loads may become an issue for large-scale data sets.

\textbf{Contributions:} We propose a general, variational and bound-optimization framework of fair clustering, which integrates an original Kullback-Leibler (KL) fairness term with a large class of clustering objectives, including both prototype-based (e.g., K-means/K-median) and graph-based (e.g., Normalized Cut or Ratio Cut). Fundamentally different from the existing combinatorial and spectral solutions, our variational multi-term approach enables to control the trade-off levels between the fairness and clustering objectives. We derive a general tight upper bound based on a concave-convex decomposition of our fairness term, its Lipschitz-gradient property and the Pinsker's inequality. Our tight upper bound can be jointly optimized with various clustering objectives, while yielding a scalable solution, with convergence guarantee. Interestingly, at each iteration, our general variational fair-clustering algorithm performs an independent update for each assignment variable. Therefore, it can easily be distributed for large-scale datasets. This scalibility is important as it enables to explore different trade-off levels between fairness and the clustering objective. Unlike the constrained spectral relaxation in \cite{Kleindessner2019}, our formulation does not require computing its eigenvalue decomposition. We report comprehensive evaluations and comparisons with state-of-the-art methods over various fair-clustering benchmarks, which show that our variational method can yield highly competitive solutions in terms of fairness and clustering objectives, while being scalable and flexible.

\section{Proposed formulation}
Let $\XX =\{ \mathbf{x}_p \in \mathbb{R}^M, p = 1 , \dots, N\}$ denote a set of $N$ data points to be assigned 
to $K$ clusters, and $\Sbold$ is a soft cluster-assignment vector: $\Sbold = [\sbold_1, \dots ,\sbold_N]\in [0,1]^{NK}$.  For each point $p$, $\sbold_p = [s_{p,k}] \in [0, 1]^K$ is the probability simplex vector verifying $\sum_k s_{p,k} = 1$. 
Suppose that the data set contains $J$ different demographic groups, with vector $V_j=[v_{j,p}]\in\{0,1\}^N$ indicating point assignment to group $j$: $v_{p, j}=1$ if data point $p$ is in group $j$ and $0$ otherwise. 
We propose the following general variational formulation for optimizing any clustering objective ${\cal F}(\Sbold)$
with a fairness penalty, while constraining each $\sbold_p$ within the  
$K$-dimensional probability simplex $\nabla_K = \{\yy \in [0, 1]^K \; | \; {\mathbf 1}^t \yy = 1 \}$: 
\begin{equation}
\min_{\Sbold}{\cal F}(\Sbold) + \lambda \sum_k \mathcal{D}_{\mbox{\tiny KL}}(U||P_k) \quad \text{s.t.} \quad \sbold_p  \in \nabla_K \; \forall p
\label{eq:fair_cl_2}
\end{equation}
$\mathcal{D}_{\mbox{\tiny KL}}(U||P_k)$ denotes the Kullback-Leibler (KL) divergence between the given (required) demographic proportions $U = [\mu_j]$ and the marginal probabilities of the demographics within cluster $k$:
\begin{equation}
\label{eq:Pk}
P_k =[P(j|k)]; \; P(j|k) = \frac{V_j^tS_k}{\one^tS_k} \forall j,
\end{equation}
where $S_k=[s_{p,k}] \in[0,1]^N$ is the $N$-dimensional vector \footnote{The set of $N$-dimensional vectors $S_k$ and the set of simplex vectors $\sbold_p$ are two equivalent ways for representing assignment variables. However, we use $S_k$ here for a clearer presentation of the problem, whereas, as will be clearer later, simplex vectors $\sbold_p$ will be more convenient in the subsequent optimization part.} containing point assignments to cluster $k$, and $t$ denotes the transpose operator.
Notice that, at the vertices of the simplex (i.e., for hard binary assignments), $V_j^tS_k$ counts the number of points within the intersection of demographic $j$ and cluster $k$, whereas $\one^tS_k$ is the total number of points within cluster $k$.  
Parameter $\lambda$ controls the trade-off between the clustering objective and fairness penalty. The problem in \eqref{eq:fair_cl_2} is challenging due to the ratios of summations in the fairness penalty and the simplex constraints. 
Expanding KL term $\mathcal{D}_{\mbox{\tiny KL}}(U||P_k)$ and discarding constant $\mu_j\log \mu_j$, our objective in \eqref{eq:fair_cl_2} becomes equivalent to minimizing the following functional with respect to the relaxed assignment variables, and subject to the simplex constraints:
\begin{equation}
{\cal E}(\Sbold) = \underbrace{{\cal F}(\Sbold)}_{\text{clustering}} + \lambda \underbrace{\sum_k\sum_j-\mu_j\log P(j|k)}_{\text{fairness}}
\label{eq:fair_cl_3}
\end{equation}
Observe that, in Eq. \eqref{eq:fair_cl_3}, the fairness penalty becomes a cross-entropy between the given (target) proportion $U$ and the marginal probabilities $P_k$ of the demographics within cluster $k$. 
Notice that our fairness penalty decomposes into convex and concave parts: 
\begin{equation}
\label{concave-convex-decomposition}
-\mu_j\log P(j|k) = \underbrace{\mu_j\log \one^tS_k}_{\text{concave}} \underbrace{- \mu_j\log V_j^tS_k}_{\text{convex}}
\end{equation}
This enables us to derive the following tight bounds (auxiliary functions) for minimizing our general fair-clustering model in \eqref{eq:fair_cl_3} using a quadratic bound and Lipschitz-gradient property of the convex part, along with Pinsker's inequality, and a first-order bound on the concave part. 

\begin{Defn} ${\cal A}_i(\Sbold) $ is an {\em auxiliary function} of objective ${\cal E}(\Sbold)$ if it is a tight upper bound at current solution $\Sbold^i$, i.e.,
it satisfies the following conditions:
\begin{subequations}
\begin{align}
{\cal E}(\Sbold) \, &\leq \, {\cal A}_i(\Sbold), \, \forall \Sbold \label{general-second-aux} \\
{\cal E}(\Sbold^i) &= {\cal A}_i(\Sbold^i) \label{general-third-aux} 
\end{align}
\label{Eq:Auxiliary_function_conditions}
\end{subequations}
where $i$ is the iteration index.
\end{Defn}

Bound optimizers, also commonly referred to as Majorize-Minimize (MM) algorithms \cite{Zhang2007}, update the current solution $\Sbold^i$ to the 
next by optimizing the auxiliary function:
\begin{equation}
\label{Aux_function_optim}
\Sbold^{i+1} = \arg \min_{\Sbold} {\cal A}_i(\Sbold)
\end{equation}
When these updates correspond to the global optima of the auxiliary functions, MM procedures enjoy a strong guarantee: The original objective function ${\cal E}(\Sbold)$ 
does not increase at each iteration:
\begin{equation}
\label{monotonicity-condition}
{\cal E}(\Sbold^{i+1}) \leq {\cal A}_i(\Sbold^{i+1}) \leq {\cal A}_i(\Sbold^i) = {\cal E}(\Sbold^{i})
\end{equation}
This general principle is widely used in machine learning as it transforms a difficult problem into a sequence of easier sub-problems \cite{Zhang2007}. 
Examples of well-known bound optimizers include concave-convex procedures (CCCP) \cite{Yuille2001}, expectation maximization (EM) algorithms and 
submodular-supermodular procedures (SSP) \cite{Narasimhan2005}, among others. The main technical difficulty in bound optimization is how to derive an auxiliary function. In the following, we derive auxiliary functions for our general fair-clustering 
objectives in \eqref{eq:fair_cl_3}.  

\begin{Prop}[Bound on the fairness penalty] Given current clustering solution $\Sbold^i$ at iteration $i$, we have the following auxiliary function on the fairness term in \eqref{eq:fair_cl_3}, up to additive and multiplicative constants, and for current solutions in which each demographic is represented by at least one point in each cluster (i.e., $V_j^t S_k^i \geq 1 \, \forall \, j,k$):
\begin{eqnarray}
{\cal G}_i(\Sbold)& \propto \sum_{p =1}^{N} \sbold_p^t(\bbold_p^i + \log \sbold_p - \log \sbold_p^i) \nonumber \\
\mbox{\em with}&\bbold_p^i = [b_{p,1}^i, \dots,b_{p,K}^i] \nonumber \\
& b_{p,k}^i= \frac{1}{L}\mathlarger{\sum}_j \left ( \frac{\mu_j}{\one^t S_k^i} - \frac{\mu_j v_{j,p}}{V_j^t S_k^i} \right )
\label{Aux-function-fairness}
\end{eqnarray}
where $L$ is some positive Lipschitz-gradient constant verifying $L \leq N$.
\end{Prop}
{\em Proof:} We provide a detailed proof in the supplemental material. Here, we give the main technical ingredients for obtaining our bound. 
 We use a quadratic bound and a Lipschitz-gradient property for the convex part, and a first-order bound on the concave part. We further bound the quadratic distances between simplex variables with the Pinsker's inequality \cite{csiszar2011information}.
 This step avoids completely point-wise Lagrangian-dual projections and inner iterations for handling the simplex constraints, yielding scalable (parallel) updates, with convergence guarantee.
 
\begin{table*}[t]
\centering
\begin{adjustbox}{max width=\textwidth}
\begin{tabular}{P{1.5cm}|P{3.6cm}|P{4.5cm}|P{5.5cm}}
\toprule
\textbf{Clustering} & ${\cal F}(\Sbold)$ &$\abold_p^i=[a_{p,k}^i],\; \forall k$& \textbf{Where}\\
\toprule
K-means & $\sum_N\sum_ks_{p,k}(\xx_p - \mathbf{c}_k)^2$ &$a_{p,k}^i = (\xx_p - \mathbf{c}_k^i)^2$&$\mathbf{c}_k^i = \frac{\XX^t S_k^i}{\one^t S_k^i}$ \\
\midrule
K-median  & $\sum_N \sum_k s_{p,k}\mathtt{d}(\xx_p, \mathbf{c}_k)$ & $a_{p,k}^i = \mathtt{d}(\xx_p,\mathbf{c}_k^i)$ & $\mathbf{c}_k^i = \underset{p \neq q}{\argmin}~\mathtt{d} (\xx_p,\xx_q)$, \newline $\mathtt{d}$ is a distance metric \\
\midrule
Ncut & 
$K - \sum_k \frac{S_k^t\mathbf{W}S_k}{\mathbf{d}^t S_k}$ &$a_{p,k}^i = d_p z_k^i -\frac{2\sum_q w(\xx_p,\xx_q)s_{p,k}^i}{\mathbf{d}^t S_k^i}$ & $z_k^i =\frac{(S_k^i)^t\mathbf{W}S_k^i}{(\mathbf{d}^t S_k^i)^2}$\newline $\mathbf{d} = [d_p]$, with $d_p= \sum_q w(\xx_p,\xx_q);\forall p$ 
\newline $\mathbf{W} = [w(\xx_p,\xx_q)]$ is an affinity matrix \\
\bottomrule
\end{tabular}
\end{adjustbox}
\caption{Auxiliary functions of several well-known clustering objectives. 
}
\label{tab:a_p}
\end{table*}

\begin{Prop}[Bound on the clustering objective] Given current clustering solution $\Sbold^i$ at iteration $i$, the auxiliary functions for several 
popular clustering objectives ${\cal F}(\Sbold)$ take the following general form:  
\begin{eqnarray}
&{\cal H}_i(\Sbold)=\sum_{p =1}^{N} \sbold_p^t \abold_p^i
\label{Aux-function-clustering}
\end{eqnarray}
where point-wise (unary) potentials $\abold_p^i$ are given in Table~\ref{tab:a_p}.
\end{Prop} 
{\em Proofs:} We give detailed proofs in the supplemental material. Here, we provide the main technical aspects: For the Ncut objective, the derivation of the auxiliary function is based on the fact that, for positive semi-definite affinity matrix $\mathbf{W}$, the Ncut objective is concave \cite{Tang2019KernelCK}. Therefore, the first-order approximation at the current solution is an auxiliary function. For the prototype-based objectives, deriving an auxiliary function follows from the observation that the optimal parameters $\mathbf{c}_k$, i.e., those that minimize the objective in closed-form, correspond to the sample means/medians within the clusters. These auxiliary functions correspond to the standard K-means and K-median procedures, which can be viewed as bound optimizers \cite{Tang2019KernelCK}. 

\begin{Prop}[Bound on the fair-clustering functional]
Given current clustering solution $\Sbold^i$, 
at iteration $i$, and bringing back the trade-off parameter $\lambda$, we have the following auxiliary function for the general fair-clustering objective ${\cal E}(\Sbold)$ in Eq. \eqref{eq:fair_cl_3}:
\begin{eqnarray}
&{\cal A}_i(\Sbold)=\sum_{p =1}^{N} \sbold_p^t( \abold_p^i + \lambda \bbold_p^i + \log \sbold_p - \log \sbold_p^i)
\label{Aux-function-total}
\end{eqnarray}
\end{Prop}
{\em Proof:}
It is straightforward to check that sum of auxiliary functions, each corresponding to a term in the objective, is also an auxiliary function of the 
overall objective.

Notice that, at each iteration, our auxiliary function in \eqref{Aux-function-total} is the sum of {\em independent} functions, each corresponding 
to a single data point $p$. 
Therefore, our minimization problem in \eqref{eq:fair_cl_3} can be tackled by optimizing each term over $\sbold_p$, subject to the simplex constraint, 
and independently of the other terms, while guaranteeing convergence:
\begin{equation}
\label{AUX-Form-each-variable}
\min_{\sbold_p \in \nabla_K} \sbold_p^t( \abold_p^i + \lambda \bbold_p^i + \log \sbold_p - \log \sbold_p^i), \, \forall p
\end{equation}
Also, notice that, in our derived auxiliary function, we obtained a convex negative entropy barrier function $\sbold_p \log \sbold_p$, which comes from the convex part in our fairness penalty. 
This entropy term is very interesting as it avoids completely expensive projection steps and Lagrangian-dual inner iterations for the simplex constraint of each point.
It yields closed-form updates for the dual variables of constraints $\mathbf{1}^t\sbold_p = 1$ and restricts the domain of each $\sbold_p$ to non-negative values, avoiding 
extra dual variables for constraints $\sbold_p \geq0$. Interestingly, entropy-based barriers are commonly used in Bregman-proximal optimization \cite{Yuan2017}, and 
have well-known computational benefits when handling difficult simplex constraints \cite{Yuan2017}. However, they are not very common in the general context of clustering. 

The objective in \eqref{AUX-Form-each-variable} is the sum of convex functions with affine simplex constraints $\one^t\sbold_p=1$. As strong duality holds for the convex objective and the affine simplex constraints, the solutions of the Karush-Kuhn-Tucker (KKT) conditions minimize globally the auxiliary function. The KKT conditions yield a closed-form solution for both primal variables $\sbold_p$ and the dual variables (Lagrange multipliers) corresponding to simplex constraints $\one^t\sbold_p=1$. 
\begin{equation}
\label{chap3:soft-max-updates-all-variables-vector}
\sbold_{p}^{i+1} = \frac{\sbold_p^i \exp (-({\mathbf a}_p^i+ \lambda{\mathbf b}_p^i)) }{{\mathbf 1}^t [\sbold_p^i\exp (-({\mathbf a}_p^i+ \lambda{\mathbf b}_p^i))]} \, \, \forall \, p 
\end{equation}

Notice that each closed-form update in \eqref{chap3:soft-max-updates-all-variables-vector}
is within the simplex. 
We give the pseudo-code of the proposed fair-clustering in \textbf{Algorithm \ref{alg}}. The algorithm can be used for any specific clustering objective, e.g., K-means or Ncut, among others, by providing the corresponding $\abold^i_p$. The algorithm consists of an inner and an outer loop.
The inner iterations updates $\sbold_p^{i+1}$ using \eqref{chap3:soft-max-updates-all-variables-vector} until ${\cal A}_i(\Sbold)$ does not change, with the clustering term $\abold^i_p$ fixed from the outer loop. The outer iteration re-computes $\abold^i_p$ from the updated $\sbold_p^{i+1}$. The time complexity of each inner iteration is $\mathcal{O}(NKJ)$. Also, the updates are independent for each data $p$ and, thus, can be efficiently computed in parallel. In the outer iteration, the time complexity of updating $\abold_p^i$ depends on the chosen clustering objective. For instance, for K-means, it is $\mathcal{O}(NKM)$, and, for Ncut, it is $\mathcal{O}(N^2K)$ for full affinity matrix $\mathbf{W}$ or much lesser for a sparse affinity matrix. Note that $\abold_p^i$ can be computed efficiently in parallel for all the clusters.

\begin{figure*}
\centering
\includegraphics[width=0.7\textwidth]{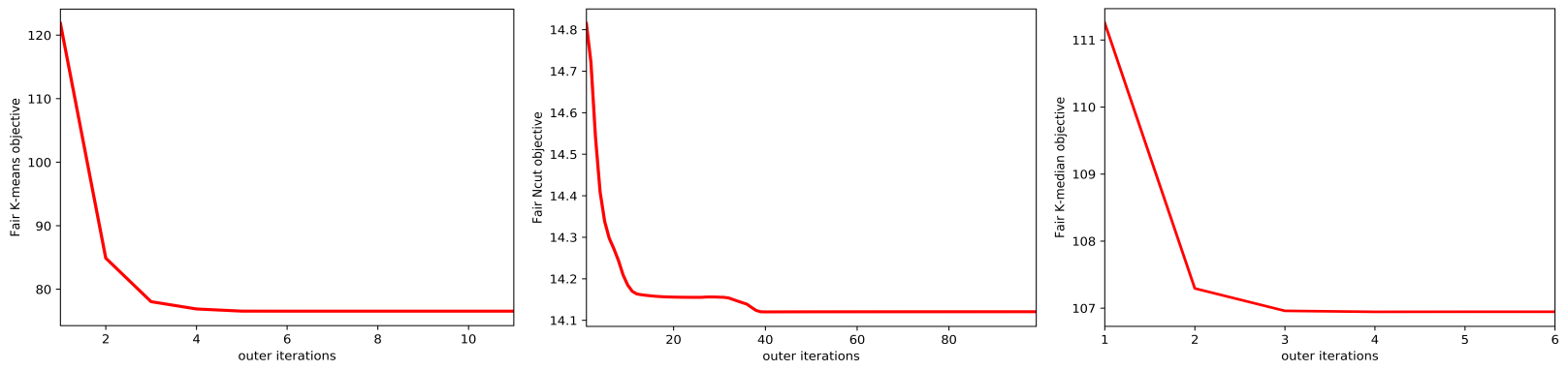}
\caption{The convergence of the proposed bound optimizers for minimizing several fair-clustering objectives in \eqref{eq:fair_cl_3}: Fair K-means, Fair Ncut and Fair K-median. The plots are based on the \textit{Synthetic} dataset.} 
\label{fig:convergence}
\end{figure*}

\begin{figure*}[htbp]
\centering
\includegraphics[width=0.9\textwidth]{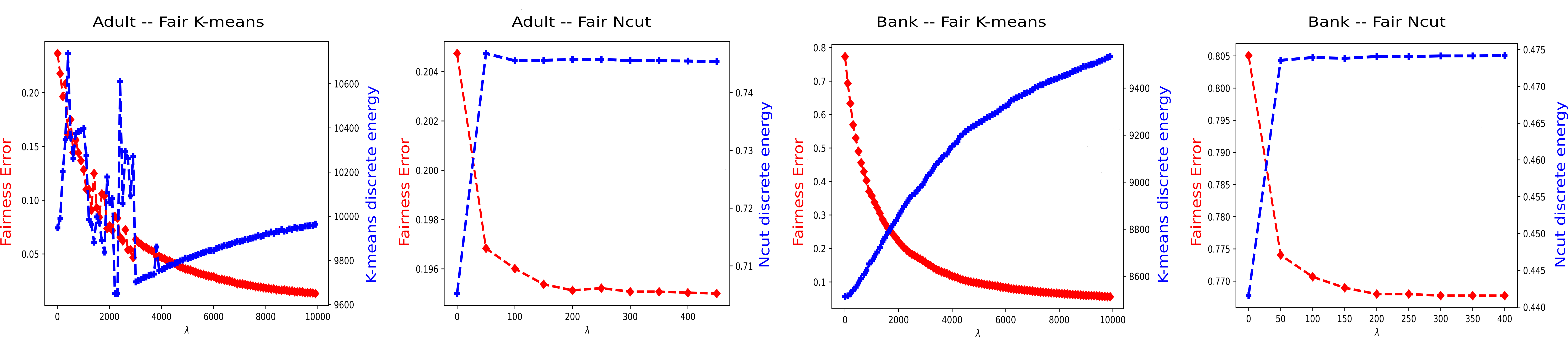}
\caption{Clustering/Fairness objectives vs. $\lambda$. }
\label{fig:lambda_vs_kl}
\end{figure*}

\textbf{Convergence and monotonicity guarantees:} Our variational model belongs to the family of MM procedures, whose theoretical guarantees are well-studied in the literature, e.g., \cite{vaida2005parameter}. In fact, the MM principle can be viewed as a generalization of well-known expectation-maximization (EM). Therefore, in general, MM algorithms inherit the monotonicity and convergence guarantees of EM algorithms, as detailed in the theoretical discussion in \cite{vaida2005parameter}. Theorem 3 in \cite{vaida2005parameter} states a condition for convergence of the general MM procedure to a local minimum: The auxiliary function has a unique global minimum, which should be obtained at each iteration when solving \eqref{Aux_function_optim}. This condition is important to guarantee, for instance, the monotonicity in \eqref{monotonicity-condition}. Our formulation satisfies this condition. In our case, the auxiliary function in \eqref{Aux-function-total} is strictly convex, as it is the sum of linear terms and a strictly convex term (the negative entropy), and is optimized under affine simplex constraints. Therefore, at each iteration, the closed-form solutions we obtained in \eqref{chap3:soft-max-updates-all-variables-vector} correspond to the unique global minimum of auxiliary function ${\cal A}_i(\Sbold)$ in \eqref{Aux-function-total}. Our plots in Fig. \ref{fig:convergence} confirm the convergence and monotonicity of our general MM procedure for several fair-clustering objectives.

\textbf{Exploring different trade-off levels via multiplier $\lambda$:} Our variational multi-term formulation enables to explore several levels of trade-off between the clustering and fairness objectives via multiplier parameter $\lambda$, unlike the existing fair-clustering methods. In practice, we run in parallel our algorithm for several values of $\lambda$ and choose the smallest value of $\lambda$ that satisfies a pre-defined level of fairness error, i.e., $\mathcal{D}_{\mbox{\tiny KL}}(U||P_k) \leq \epsilon$. This is conceptually similar to standard penalty and augmented-Lagrangian approaches in constrained optimization, where the weights of the penalties\footnote{In standard constrained optimization, penalties typically take a quadratic form, unlike our method, which is based on a KL divergence penalty.} are gradually increased, until reaching a certain pre-defined precision (or duality gap)  for the constraints; see Chapter Chapter 17.1 in \cite{nocedal2006numerical}. The difference here is that we run independently for each $\lambda$, which can be implemented in parallel. As illustrated by the plots in Fig. \ref{fig:lambda_vs_kl}, when $\lambda$ increases, the fairness error decreases and the clustering objective increases, which is intuitive. As discussed in more details below (Tables \ref{tab:tab-results-compare-1}, \ref{tab:tab-results-compare-3} and \ref{tab:tab-results-compare-2}), our variational formulation can achieve small fairness errors (competitive with the existing state-of-the-art fair-clustering methods), but with much better clustering objectives, consistently across all the data sets.
\begin{algorithm}[htbp]
   \caption{Proposed Fair-clustering}
   \label{alg}
\begin{algorithmic}
   \STATE {\bfseries Input:} $\XX$, Initial seeds, $\lambda$, $U$, $\{V_j\}_{j=1}^J$
   \STATE {\bfseries Output:} Clustering $labels \in \{1,..,K\}^N$
   \STATE Initialize $labels$ from initial seeds.
   \STATE Initialize $\Sbold$ from $labels$.
   \STATE Initialize $i = 1$.
   \REPEAT
   \STATE Compute $\abold_p^i$ from $\Sbold$ (see Table \ref{tab:a_p}).
   \STATE Initialize $\sbold^i_{p} =\frac{\exp(-\abold^i_p)}{\mathbf{1}^t\exp(-\abold^i_p)}$.
   \REPEAT
   \STATE Compute $\sbold^{i+1}_p$ using \eqref{chap3:soft-max-updates-all-variables-vector}.
   \STATE $\sbold^{i}_p \leftarrow \sbold^{i+1}_p$.
   \STATE $\Sbold = [\sbold^{i}_{p}];\; \forall p$.
   \UNTIL{${\cal A}_{i} (\Sbold)$ in \eqref{Aux-function-total} does not change}
   \STATE $i = i+1$.
   \UNTIL{${\cal E} (\Sbold)$ in \eqref{eq:fair_cl_3} does not change}
   \STATE $l_p = \underset{k}{\argmax}~s_{p,k}; \forall p$.
   \STATE $labels = \{l_p\}_{p=1}^N$.
\end{algorithmic}
\end{algorithm}
\begin{table*}[!t]
\centering
\small
\begin{tabular}{|l|c|c|c|c|}
\hline
\multirow{2}{*}{\textbf{Datasets}}&\multicolumn{4}{c|}{\textbf{Fair K-median}}\\ 
\cline{2-5}
&\multicolumn{2}{c|}{Objective}& \multicolumn{2}{c|}{fairness error / balance}\\
\cline{2-5}  
&Backurs et al. 2019 & Ours & Backurs et al. 2019 & Ours \\
\hline
Synthetic ($N=400,~J=2,~\lambda=10$) &299.859&\textbf{292.4}&\textbf{0.00}/\textbf{1.00}&\textbf{0.00}/\textbf{1.00}\\
Synthetic-unequal ($N=400,~J=2,~\lambda=10$) &185.47&\textbf{174.81}&0.77/0.21&\textbf{0.00}/\textbf{0.33}\\
Adult ($N=32,561,~J=2,,~\lambda=9000$) &19330.93&\textbf{18467.75}&0.27/0.31&\textbf{0.01}/\textbf{0.43}\\
Bank ($N=41,108,~J=3,~\lambda=9000$) &N/A&\textbf{19527.08}&N/A&\textbf{0.02}/\textbf{0.18}\\
Census II ($N=2,458,285,~J=2,~\lambda=500000$) &2385997.92&\textbf{1754109.46}&0.41/0.38&\textbf{0.02}/\textbf{0.78}\\
\hline
\end{tabular}
\caption{Comparison of the proposed Fair K-median to \cite{backurs2019scalable}.}
\label{tab:tab-results-compare-1}
\end{table*}

\begin{table*}[!h]
\centering
\small
\begin{tabular}{|l|c|c|c|c|}
\hline
\multirow{3}{*}{\textbf{Datasets}}&\multicolumn{4}{c|}{\textbf{Fair K-means}}\\ 
\cline{2-5}
&\multicolumn{2}{c|}{Objective}& \multicolumn{2}{c|}{fairness error / balance}\\
\cline{2-5}  
&Bera et al. 2019 & Ours & Bera et al. 2019 & Ours \\
\hline
Synthetic ($N=400,~J=2,~\lambda=10$) & 758.43&\textbf{207.80}& \textbf{0.00 / 1.00} & \textbf{0.00 / 1.00}\\
Synthetic-unequal ($N=400,~J=2,~\lambda=10$) &180.00 & \textbf{159.75} & \textbf{0.00 / 0.33} & \textbf{0.00 / 0.33}\\
Adult ($N=32,561,~J=2,~\lambda=9000$) &10913.84 &\textbf{9984.01} & \textbf{0.018} / \textbf{0.41} & \textbf{0.018} / \textbf{0.41}\\
Bank ($N=41,108,~J=3,~\lambda=6000$) & 11331.51 & \textbf{9392.20} & \textbf{0.03} / 0.16 & 0.05 / \textbf{0.17}\\
Census II ($N=2,458,285,~J=2,~\lambda=500000$)  &1355457.02&\textbf{1018996.53}&0.07/0.77&\textbf{0.02/0.78}\\
\hline
\end{tabular}
\caption{Comparison of the proposed Fair K-means to \cite{bera2019fair}.}
\label{tab:tab-results-compare-3}
\end{table*}

\begin{table*}[t]
\centering
\small
\begin{tabular}{|l|c|c|c|c|}
\hline
\multirow{3}{*}{\textbf{Datasets}}&\multicolumn{4}{c|}{\textbf{Fair NCUT}}\\ 
\cline{2-5}
&\multicolumn{2}{c|}{Objective}& \multicolumn{2}{c|}{fairness error / balance}\\
\cline{2-5}  
&Kleindessner et al. 2019 & Ours & Kleindessner et al. 2019 & Ours \\
\hline
Synthetic ($N=400,~J=2,~\lambda=10$) &\textbf{0.0}&\textbf{0.0}&0.00/1.00&\textbf{0.0}/\textbf{1.00}\\
Synthetic-unequal ($N=400,~J=2,~\lambda=10$) &\textbf{0.03}&0.06&\textbf{0.00}/\textbf{0.33}&\textbf{0.00}/\textbf{0.33}\\
Adult ($N=32,561,~J=2,~\lambda=10$) &\textbf{0.47}&0.74&\textbf{0.06}/\textbf{0.32}&0.08/0.30\\
Bank ($N=41,108,~J=3,~\lambda=40$) &N/A&\textbf{0.58}&N/A&\textbf{0.39}/\textbf{0.14}\\
Census II ($N=2,458,285,~J=2,~\lambda=100$)  &N/A&\textbf{0.52}&N/A&\textbf{0.41/0.43}\\
\hline
\end{tabular}
\caption{Comparison of the proposed Fair NCut to \cite{Kleindessner2019}.}
\label{tab:tab-results-compare-2}
\end{table*}

\section{Experiments}
In this section, we present comprehensive empirical evaluations of the proposed fair-clustering algorithm, along with comparisons with state-of-the-art fair-clustering techniques. We choose three well-known clustering objectives: K-means, K-median and Normalized cut (Ncut), and integrate our fairness-penalty bound with the corresponding clustering bounds $\abold_p$ (see Table \ref{tab:a_p}). We refer to our bound-optimization versions as: Fair K-means, Fair K-median and Fair Ncut. Note that our formulation can be used for other clustering objectives (if a bound could be derived for the objective).

We investigate the effect of fairness on the original discrete (i.e., w.r.t. binary assignment variables) clustering objectives, and compare with the existing methods. We evaluate the results in terms of the balance of each cluster $S_k$ in \eqref{cluster-balance}, and define the overall \textbf{balance} of the clustering as $\mbox{balance} = \min_{S_k} \mbox{balance}(S_k)$. We further propose to evaluate the \textbf{fairness error}, which is the KL divergence $\mathcal{D}_{\mbox{\tiny KL}}(U||P_k)$ in \eqref{eq:fair_cl_2}. This KL measure becomes equal to zero when the proportions of the demographic groups within all the output clusters match the target distribution.  For Ncut, we use $20$-nearest neighbor affinity matrix, $\mathbf{W}$: $w(\xx_p, \xx_q) = 1$ if data point $\xx_q$ is within the $20$-nearest neighbors of $\xx_p$, and equal to $0$ otherwise. In all the experiments, we fixed $L=2$ and found that this does not increase the objective (see the detailed explanation in the supplemental material). We standardize each dataset by making each feature attribute to have zero mean and unit variance. We then performed L2-normalization of the features, and used the standard K-means++ \cite{arthur2007k} to generate initial partitions for all the models.

\subsection{Datasets}

\textbf{Synthetic datasets. } We created two types of synthetic datasets according to the proportions of the demographics, each having two clusters and a total of $400$ data points in 2D features (figures in the supplemental material). The \textit{Synthetic} dataset contains two perfectly balanced demographic groups, each having an equal number of $200$ points. For this data set, we imposed target target proportions $U = [0.5, 0.5]$. To experiment with our fairness penalty with unequal proportions, we also used \textit{Synthetic-unequal} dataset with 300 and 100 points within each of the two demographic groups. In this case, we imposed target proportions $U = [0.75, 0.25]$.

\textbf{Real datasets.}
We use three datasets from the UCI machine learning repository \cite{Dua:2019}, one large-scale data set whose demographics are 
balanced (Census), along with two other data sets with various demographic proportions: 

\textit{Bank} \footnote{https://archive.ics.uci.edu/ml/datasets/Bank+Marketing} dataset contains $41188$ number of records of direct marketing campaigns of a Portuguese banking institution corresponding to each client contacted \cite{moro2014data}. Note that, the previous fair clustering methods \cite{bera2019fair,backurs2019scalable} used a much smaller version of Bank dataset with only $4520$ number of records with $J=2$ and $3$ attributes. Instead, we utilize the marital status as the sensitive attribute, which contains three groups ($J=3$) -- single, married and divorced -- and removed the `'Unknown'' marital status. Thus, we have $41,108$ records in total. We chose $6$ numeric attributes (age, duration, euribor of 3 month rate, no. of employees, consumer price index and number of contacts performed during the campaign) as features. We set the number of clusters $K =10$, and impose the target proportions of three groups $U=[0.28,0.61,0.11]$ within each cluster.

\textit{Adult}\footnote{https://archive.is.uci/ml/datasets/adult} is a US census record data set from 1994. The dataset contains $32,561$ records. We used the gender status as the sensitive attribute, which contains $10771$ females and $21790$ males. We chose the $5$ numeric attributes as features, set the number of clusters to $K =10$, and impose 
proportions $U=[0.33,0.67]$ within each cluster.

\textit{Census}\footnote{https://archive.ics.uci.edu/ml/datasets/US+Census+Data+(1990)} is a large-scale data set corresponding to a US census record data from 1990. 
The dataset contains $2,458,285$ records. We used the gender status as the sensitive attribute, which contains $1,191,601$ females and $1,266,684$ males. We chose the $25$ numeric attributes as features, similarly to \cite{backurs2019scalable}. We set the number of clusters to $K =20$, and imposed proportions $U=[0.48,0.52]$ within each cluster.


 \begin{figure*}[htbp]
\centering
\includegraphics[width=0.65\textwidth]{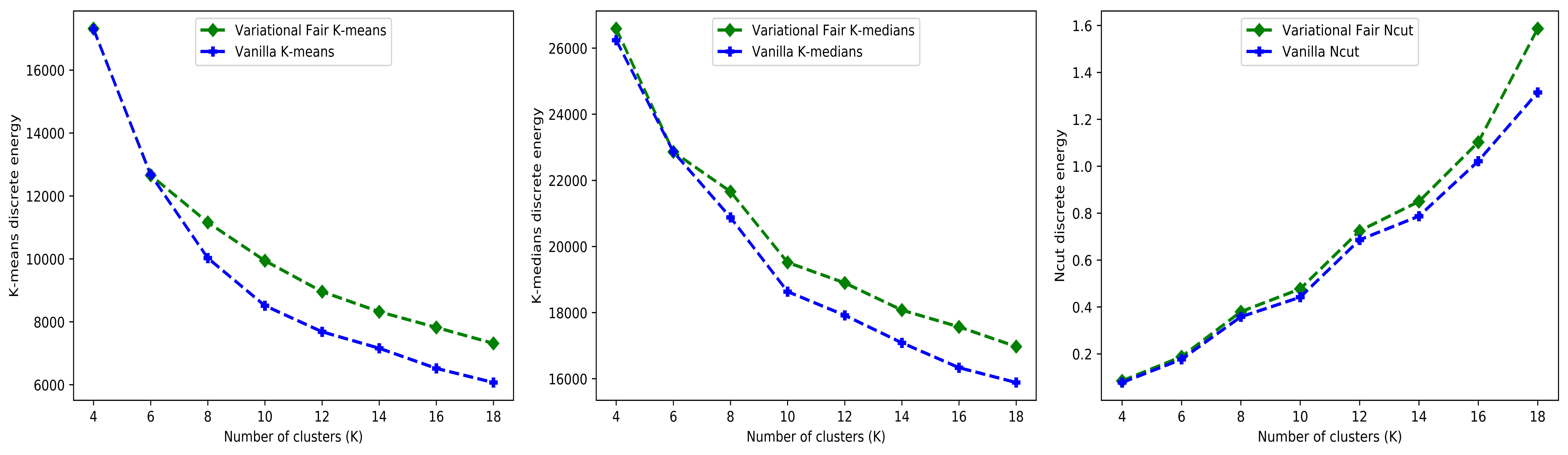}
\caption{Effect of $K$ on the clustering objectives for vanilla clustering and our variational fair clustering methods, including the K-means, K-medians and Ncut objectives. The results are shown for the Bank dataset. Note that, for each plot, the value of multiplier $\lambda$ is fixed.
}
\label{fig:K_vs_objective}
\end{figure*}

 \subsection{Results}
 In this section, we discuss the results of the different experiments to evaluate the proposed general variational framework for Fair K-means, Fair K-median and Fair Ncut. 
 We further report comparisons with \cite{bera2019fair}, \cite{backurs2019scalable} and \cite{Kleindessner2019} in terms of discrete fairness measures and clustering objectives.

 \textbf{Trade-off between clustering and fairness objectives:} We assess the effect of imposing fairness constraints on the original clustering objectives. In each plot in Fig. \ref{fig:lambda_vs_kl}, the blue curve depicts the discrete-valued clustering objective ${\cal F}(\Sbold)$ (K-means or Ncut) obtained at convergence as a function of $\lambda$. The fairness error is depicted in red. Observe that, when multiplier $\lambda$ increases (starting from a certain value), the discrete clustering objective increases while the fairness error decreases, which is intuitive. Also, the fairness error approaches $0$ when $\lambda \rightarrow +\infty$, and both the clustering and fairness objectives tend to reach a plateau starting from a certain value of $\lambda$. The scalability of our model is highly relevant because it enables us to explore several solutions, each corresponding to a different value of multiplier $\lambda$, and to choose the smallest $\lambda$ (i.e., the best clustering objective) that satisfies a pre-defined fairness level $\mathcal{D}_{\mbox{\tiny KL}}(U||P_k) \leq \epsilon$. As detailed below, this flexibility enabled us to obtain better solutions, in terms of fairness and clustering objectives, than several recent fair-clustering methods. Low fairness errors are typically achieved with large values of $\lambda$. This is due to the fact that the scale of the fairness penalty could be much smaller than the clustering objectives. Notice that, for relatively small values of $\lambda$, the K-means objective (blue curve) for the Adult dataset has an oscillating behaviour. This might be due to the fact that, for small $\lambda$, the K-means objective dominates the KL fairness term. However, after a certain value of $\lambda$ ($\lambda \geq 4000$), the curves become smooth, with a predictable behaviour (i.e., the fairness term decreases and the clustering term increases). When the clustering objective dominates, the oscillating behaviour might be due to the local minima of bound optimization for the K-means term. We hypothesize that, with higher values of $\lambda$, the KL fairness term ``convexifies" the function, and facilitates optimization. With smaller values of $\lambda$, the K-means term dominates, with possibilities of being stuck in local minima (K-means is well-known to be sensitive to the initial conditions).
 
 \textbf{Clustering cost with respect to $K$:} Fig. \ref{fig:K_vs_objective} depicts how the clustering objectives behave w.r.t to the number of clusters $K$, with and without the fairness constraints. We plot the discrete clustering objective vs. $K$ for K-means, K-medians and Ncut, using the Bank dataset, with each plot corresponding to a fixed multiplier $\lambda$. In both cases (i.e., with and without the fairness constraints), the obtained clustering objectives decrease with $K$, with the gap between the clustering objective obtained under fairness constraints and the vanilla clustering increasing with $K$. Those experimental observations are consistent with the observations in \cite{bera2019fair}.

\textbf{Comparisons to state-of-the-art methods:} Our algorithm is flexible as it can be used
 in conjunction with different well-known clustering objectives. This enabled us to compare our Fair K-median, Fair K-means and Fair Ncut versions 
 to \cite{backurs2019scalable}, \cite{bera2019fair} and \cite{Kleindessner2019}, respectively. Tables \ref{tab:tab-results-compare-1}, \ref{tab:tab-results-compare-3} and \ref{tab:tab-results-compare-2} report comparisons in terms of the original clustering objectives, achieved minimum balances and fairness errors, for Fair K-medians, Fair K-means and Fair NCut. For our model, we run the algorithm for several values of $\lambda$ in ascending order, and choose the smallest $\lambda$ that satisfies a pre-defined level of fairness error. This flexibility and scalability enabled us to obtain significantly better clustering objectives and fairness/minimum-balance measures in comparisons to \cite{backurs2019scalable}; See Table \ref{tab:tab-results-compare-1}. It is worth noting that, for the \textit{Bank} dataset, we were unable to run \cite{backurs2019scalable} as the 
 number of demographic group is $3$ (i.e. $J>2$). 
 In comparison to \cite{bera2019fair}, our variational method achieves significantly better K-means clustering objectives, with approximately the same fairness levels. Note that, we can obtain better fairness with larger $\lambda$ values. These results highlight the benefits of our proposed variational formulation, which provides control over the trade-off between the fairness level and clustering objective.
 In the case of fair NCut, \cite{Kleindessner2019} achieved slightly better Ncut objectives than our model, while achieving similar fairness levels. However, we were unable to run the spectral solution of \cite{Kleindessner2019} for large-scale \textit{Census II} data set, and for \textit{Bank}, due to its computational and memory load (as it requires computing the eigen values of the square affinity matrix). 

 Our algorithm scales up to more than two demographic groups, i.e. when $J>2$ (e.g. \textit{Bank}), unlike many of the existing approaches. Furthermore, for NCut graph clustering, our bound optimizer can deal with large-scale data sets, unlike \cite{Kleindessner2019}, which requires eigen decomposition for large affinity matrices. Finally, the parallel structure of our algorithm within each iteration (i.e., independent updates for each assignment variable) enables to explore different values of $\lambda$, thereby choosing the best trade-off between the clustering objective and fairness error.


\newpage

\textbf{Broader Impact}: This paper deals with ensuring fairness criteria in clustering decisions, so as to avoid unfair treatment of minority groups pertaining to a sensitive attribute such as race, gender, etc. The paper is an endeavor to present a flexible mechanism, so as to relatively control the required fairness, while ensuring clustering quality at the same time. 
\appendix

\section{Proof of Proposition 1}
We present a detailed proof of \textbf{Proposition 1 (Bound on fairness)} in the paper. Recall that, in the paper, we wrote the fairness clustering problem in the following form:  
\begin{equation*}
{\cal E}(\Sbold) = \underbrace{{\cal F}(\Sbold)}_{\text{clustering}} + \lambda \underbrace{\sum_k\sum_j-\mu_j\log P(j|k)}_{\text{fairness}}
\label{eq:fair_cl_3-a}
\end{equation*}

The proposition for the bound on the fairness penalty states the following:
Given current clustering solution $\Sbold^i$ at iteration $i$, we have the following tight upper bound (auxiliary function) on the fairness term in \eqref{eq:fair_cl_3-a}, up to additive and multiplicative constants, and for current solutions in which each demographic is represented by at least one point in each cluster (i.e., $V_j^tS_k^i \geq 1 \, \forall \, j,k$):
\begin{eqnarray}
{\cal G}_i(\Sbold)& \propto \sum_{p =1}^{N} \sbold_p^t(\bbold_p^i + \log \sbold_p - \log \sbold_p^i)\nonumber\\
\mbox{\em with}&\bbold_p^i = [b_{p,1}^i, \dots,b_{p,K}^i] \nonumber\\
&b_{p,k}^i= \frac{1}{L}\mathlarger{\sum_{j}} \left ( \frac{\mu_j}{\one^tS_k^i} - \frac{\mu_jv_{j,p}}{V_j^tS_k^i} \right )
\label{Aux-function-fairness-a}
\nonumber
\end{eqnarray}
where $L$ is some positive Lipschitz-gradient constant verifying $L \leq N$

{\em Proof:}
We can expand each term in the fairness penalty in \eqref{eq:fair_cl_3-a}, and write it as the sum of two functions, one is convex and the other is concave:  
\begin{eqnarray}
-\mu_j\log P(j|k) &=& \mu_j\log \one^tS_k - \mu_j\log V_j^tS_k \nonumber\\
&=& g_1(S_k) + g_2(S_k)
\label{eq:fairness_expand}
\end{eqnarray}
Let us represent the $N\times K$ matrix $\Sbold = \{S_1, \dots ,S_K\}$ in its equivalent vector form $\Sbold = [\sbold_1, \dots ,\sbold_N]\in [0,1]^{NK}$, where $\sbold_p = [s_{p,1}, \dots, s_{p,K}]\in[0,1]^K$ is the probability simplex assignment vector for point $p$. As we shall see later, this equivalent simplex-variable representation will be convenient for deriving our bound. 

\textbf{Bound on $\tilde{g}_1(\Sbold) = \sum_k g_1(S_k)$:} \\
For concave part $g_1$, we can get a tight upper bound (auxiliary function) by its first-order approximation at current solution $S_k^i$:
\begin{eqnarray}
g_1(S_k) &\leq& g_1(S_k^i) + [\nabla g_1(S_k^i)]^t (S_k - S_k^i) \nonumber \\
&=& [\nabla g_1(S_k^i)]^t S_k + const
\label{eq:bound_1_1}
\end{eqnarray}
where gradient vector $\nabla g_1(S_k^i) =  \frac{\mu_j}{\one^tS_k^i}\one$ and $const$ is the sum of all the constant terms.
Now consider $N\times K$ matrix ${\mathbf T}_1 = \{\nabla g_1(S_1^i), \dots \nabla g_1(S_K^i)\}$ and it equivalent vector representation 
${\mathbf T}_1 = [{\mathbf t}^1_1, \dots ,{\mathbf t}^N_1]\in {\mathbb R}^{NK}$, which concatenates rows ${\mathbf t}^p_1 \in {\mathbb R}^{K}$, $p \in \{1, \dots N \}$, of the $N\times K$ matrix into a single $NK$-dimensional vector. Summing the bounds in \eqref{eq:bound_1_1} over $k \in \{1, \dots K \}$ and using the 
$NK$-dimensional vector representation of both $\Sbold$ and ${\mathbf T}_1$, we get: 
\begin{eqnarray}
\tilde{g}_1(\Sbold) &\leq& {\mathbf T}_1^t \Sbold + const
\label{eq:bound_1_2}
\end{eqnarray}

\textbf{Bound on $\tilde{g}_2(\Sbold) = \sum_k g_2(S_k)$:} \\
For convex part $g_2$, a quadratic upper bound can be found by using \textbf{Lemma 1} and \textbf{Definition 1} (both detailed at the end of the document):
\begin{eqnarray}
g_2(S_k) &\leq& g_2(S_k^i) + [\nabla g_2(S_k^i)]^t (S_k - S_k^i) + L \|S_k - S_k^i \|^2 \nonumber \\
&=& [\nabla g_2(S_k^i)]^t S_k + L\|S_k - S_k^i \|^2 + const
\label{eq:bound_2_1}
\end{eqnarray}
where gradient vector $\nabla g_2(S_k^i) = - \frac{\mu_jV_j}{V_j^tS_k^i} \in \mathbb{R}^N$ and $L$ is a valid Lipschitz constant for the gradient of $g_2$.
Similarly to earlier, consider $N\times K$ matrix ${\mathbf T}_2 = \{\nabla g_2(S_1^i), \dots \nabla g_2(S_K^i)\}$ and its equivalent vector 
representation ${\mathbf T}_2 = [{\mathbf t}^1_2, \dots ,{\mathbf t}^N_2]\in {\mathbb R}^{NK}$. Using this equivalent vector representations for matrices ${\mathbf T}_2$, $\Sbold$ and $\Sbold^i$, and summing the bounds in \eqref{eq:bound_2_1} over $k$, we get:
\begin{eqnarray}
\tilde{g}_2(\Sbold) &\leq& {\mathbf T}_2^t \Sbold + L \|\Sbold - \Sbold^i \|^2 + const
\label{eq:bound_2_2}
\end{eqnarray}
In our case, the Lipschitz constant is: $L = \sigma_{max}$, where $\sigma_{max}$ is the maximum eigen value of the Hessian matrix:
\[\nabla^2(g_2(S_k^i)) = \frac{\mu_j}{(V_j^tS_k^i)^2}V_jV_j^t.\] 

Note that, $ \|\Sbold - \Sbold^i \|^2$ is defined over the simplex variable of each data point $\sbold_p$. Thus, we can utilize \textbf{Lemma 2 (Pinsker inequality)}, which yields the following bound on $\tilde{g}_2(\Sbold)$ (Lemma 2 and its proof are detailed below):

\begin{eqnarray}
\tilde{g}_2(\Sbold) &\leq& \Sbold^t[{\mathbf T}_2+ L\log \Sbold - L\log \Sbold^i]
\label{eq:bound_2_3}
\end{eqnarray}

\textbf{Total bound on the Fairness term:}\\ 
 By taking into account the sum over all the demographics $j$ and combining the bounds for $\tilde{g}_1(\Sbold)$ and $\tilde{g}_2(\Sbold)$, we get the following bound for the fairness term:
 
\begin{eqnarray}
{\cal G}_i(\Sbold)&=&\Sbold^t\left [\sum_j({\mathbf T}_1 + {\mathbf T}_2)+ L\log \Sbold - L\log \Sbold^i \right ] \nonumber \\
&\propto&\sum_{p =1}^{N} \sbold_p^t(\bbold_p^i + \log \sbold_p - \log \sbold_p^i)\nonumber\\
\mbox{\em with}&&\bbold_p^i = [b_{p,1}^i, \dots,b_{p,K}^i] \nonumber\\
&&b_{p,k}^i= \frac{1}{L}\mathlarger{\sum_{j}} \left ( \frac{\mu_j}{\one^tS_k^i} - \frac{\mu_jv_{j,p}}{V_j^tS_k^i} \right )
\label{total_bound}
\end{eqnarray}

Note that for current solutions in which each demographic is represented by at least one point in each cluster (i.e., $V_j^tS_k^i \geq 1 \, \forall \, j,k$), the maximum eigen value of the Hessian $\nabla^2(g_2(S_k^i))$ is bounded by $N$, which means $L\leq N$. Note that, in our case, typically the term $\frac{\mu_j}{(V_j^tS_k^i)^2}$ in the Hessian is much smaller than $1$. Therefore, in practice, setting a suitable positive $L<<N$ does not increase the objective. 

\begin{defn}
A convex function $f$ defined over a convex set $\Omega \in \RR^l$ is {\em L-smooth} if 
the gradient of $f$ is Lipschitz (with a Lipschitz constant $L>0$): $\|\nabla f(\xx) - \nabla f(\yy)\| \leq L . \| \xx-\yy\|$ for all $\xx, \yy \in \Omega$. Equivalently, there exists a strictly positive $L$ such that the Hessian of $f$ verifies: $\nabla^2 f(\xx) \preceq L \II$ where $\II$ is the identity matrix.  
\end{defn}

\begin{Rem}
Let $\sigma_{max} (f)$ denotes the maximum Eigen value of $\nabla^2 f(\xx)$ is a valid Lipschitz constant for the gradient of $f$ because $\nabla^2 f(\xx) \preceq  \sigma_{max} (f)\II$ 
\end{Rem}

Lipschitz gradient implies the following bound\footnote{This implies that the distance between the $f(\xx)$ and its first-order Taylor approximation at $\yy$ is between $0$ and $L. \|\xx - \yy \|^2$. Such a distance is the Bregman divergence with respect to the $l_2$ norm.} on $f(\xx)$
\begin{Lem}[Quadratic upper bound]
If $f$ is {\em L-smooth}, then we have the following quadratic upper bound:
\begin{equation}
f(\xx) \leq f(\yy) + [\nabla f(\yy)]^t (\xx-\yy)+ L. \|\xx - \yy \|^2
\end{equation}
\end{Lem}
\begin{figure*}[t]
\centering
\includegraphics[width=\textwidth]{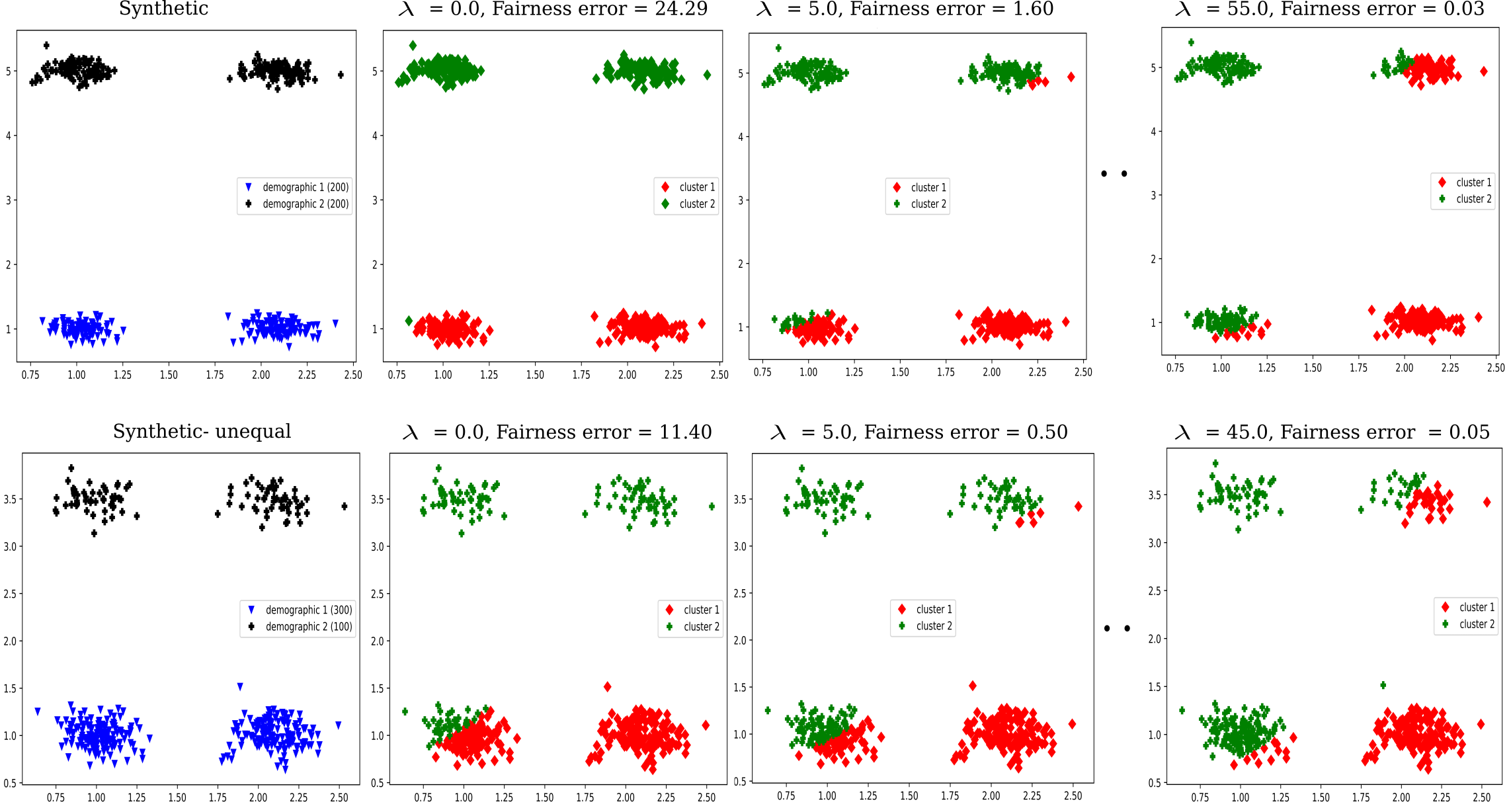}
\caption{Output clusters of Fair K-means with respect to $\lambda$ on synthetic datasets. Demographics are colored in either black or blue and the output clusters are colored in either red or green. \textbf{First row}-- 1st: \textit{Synthetic} dataset with two equal demographics. (2nd-4th): With the increased $\lambda$ parameter, the output clusters get balanced demographics. \textbf{Second row}-- 1st: \textit{Synthetic-unequal} dataset with different demographic proportions $U=[0.75,0.25]$. (2nd-4th): output clusters colored in either red or green. With the increased $\lambda$ parameter, the output clusters are according to the given proportions of demographics, with almost $0$ fairness error.}
\label{fig:lambda_vs_clusters}
\end{figure*}
{\em Proof:}
The proof of this lemma is straightforward. It suffices to start from convexity condition 
$f(\yy) \geq f(\xx) + [\nabla f(\xx)]^t(\yy-\xx)$ and use Cauchy-Schwarz inequality and the Lipschitz gradient condition:
\begin{eqnarray}
f(\xx)  &\leq& f(\yy) + [\nabla f(\xx)]^t(\xx-\yy) \nonumber \\
& = & f(\yy) + [\nabla f(\yy)]^t(\xx-\yy) \nonumber   \\ &+& [\nabla f(\xx) - \nabla f(\yy)]^t(\xx-\yy) \nonumber   \\
&\leq& 
f(\yy) + [\nabla f(\yy)]^t(\xx-\yy)  \nonumber \\ &+& \| \nabla f(\xx) - \nabla f(\yy) \|.\| \xx-\yy \| \nonumber \\
&\leq& 
f(\yy) + [\nabla f(\yy)]^t(\xx-\yy) + L. \| \xx-\yy \|^2 \nonumber
\end{eqnarray}

\begin{Lem}[Pinsker inequaltiy]
\label{KL-lower-bound}
For any $\xx$ and $\yy$ belonging to the $K$-dimensional probability simplex ${\cal S} = \{\xx \in [0, 1]^K \; | \; {\mathbf 1}^t \xx = 1 \}$, we have the following inequality:
\begin{equation}
\mathcal{D}_{\mbox{\tiny KL}}(\xx||\yy) \geq \frac{1}{2} \| \xx-\yy \|^2  
\end{equation}
where $\mathcal{D}_{\mbox{\tiny KL}}$ is the Kullback-Leibler divergence:
\begin{equation}
\mathcal{D}_{\mbox{\tiny KL}}(\xx||\yy) = \sum_k x_k \log \frac{x_k}{y_k}
\end{equation} 
\end{Lem}

{\em Proof:}
Let $q_{\oo}(\xx) = {\cal D}_{k}(\xx||\oo)$. 
The Hessian of $q_{\oo}$ is a diagonal matrix whose diagonal elements are given by: $\frac{1}{x_k}, k=1, |\dots K$. Now because $\xx \in {\cal S}$, we have $\frac{1}{x_i}>1 \quad \forall i$. Therefore, $q_{\oo}$ is $1$-{\em strongly} convex: 
$\nabla^2 q_{\oo}(\xx) \succeq \II$. This is equivalent to:
\begin{equation}
\label{l-strong-convexity}
q_{\oo}(\xx) \geq 
q_{\oo}(\yy) + [\nabla q_{\oo}(\yy)]^t (\xx-\yy) + \frac{1}{2}\| \xx-\yy \|^2
\end{equation}
The gradient of $q_{\oo}$ is given by:
\begin{equation}
\nabla q_{\oo}(\yy) = (1+\log \frac{\yy_1}{\oo_1}, \dots, 1+\log \frac{\yy_k}{\oo_k})^t.
\end{equation} 
Applying this expression to $\oo=\yy$, notice that $\nabla q_{\oo}(\yy) = \one$. Using these in expression \eqref{l-strong-convexity} for $\oo=\yy$, we get:
\begin{equation}
\mathcal{D}_{\mbox{\tiny KL}}(\xx||\yy)  \geq  \one^t(\xx-\yy) + \frac{1}{2}\| \xx-\yy \|^2
\end{equation} 
Now, because $\xx$ and $\yy$ are in ${\cal S}$, we have $\one^t(\xx-\yy) = \sum_k \xx_k - \sum_k \yy_k =1-1=0$. 
This yields the result in Lemma \ref{KL-lower-bound}.

\section{Proof of Proposition 2}

Here we present the proof of Proposition 2 [Bound on the clustering objective]: Given current clustering solution $\Sbold^i$ at iteration $i$, the auxiliary functions for several 
popular clustering objectives ${\cal F}(\Sbold)$ take the following general form:  
\begin{eqnarray}
&{\cal H}_i(\Sbold)=\sum_{p =1}^{N} \sbold_p^t \abold_p^i
\end{eqnarray}
where point-wise (unary) potentials $\abold_p^i$ are given in Table 1 in the paper.

For K-means, and for each cluster $S_k$, the sample mean of the cluster is the closed-form global optimum of summation $\sum_{p}s_{p,k}(\xx_p - \mathbf{c}_k)^2$, i.e., 
\[ \mathbf{c}_k = \argmin_{\mathbf{y}} \sum_{p}s_{p,k}(\xx_p - \mathbf{c}_k)^2 \] 
Therefore, for any $\mathbf{y}$, we have the following inequality:
\[\sum_{p}s_{p,k}(\xx_p - \mathbf{c}_k)^2 \leq \sum_{p}s_{p,k}(\xx_p - \mathbf{y})^2\]
Thus, by setting $\mathbf{y} = \mathbf{c}_k^i$ at current iteration $i$, we get the following auxiliary function for K-means:
$$
\sum_{p}s_{p,k}(\xx_p - \mathbf{c}_k)^2 \leq \sum_{p}s_{p,k}(\xx_p - \mathbf{c}_k^i)^2
$$

Similarly, we can show the same for K-median:
$$
\sum_{p}s_{p,k}\mathtt{d}(\xx_p - \mathbf{c}_k) \leq \sum_{p}s_{p,k}\mathtt{d}(\xx_p - \mathbf{c}_k^i)
$$

Thus, considering all the clusters, we write the bound on the clustering objectives in the following simplified form:
$$
{\cal H}_i(\Sbold)=\sum_{p =1}^{N} \sbold_p^t \abold_p^i
$$

Where $\sbold_p = [s_{p,1}, \ldots, s_{p,K}]$ and $\abold_p^i = [a_{p,1}^i, \ldots, a_{p,K}^i]$. In the case of K-means, we have: $a_{p,k}^i = (\xx_p - \mathbf{c}_k^i)^2$, and for K-medians, we have: $a_{p,k}^i = \mathtt{d}(\xx_p - \mathbf{c}_k^i)$.

For Ncut, the objective is: 
\[{\cal F}(\Sbold) = K - \sum_k \frac{S_k^t\mathbf{W}S_k}{\mathbf{d}^t S_k} = K- \sum_k\mathcal{F}(S_k)\] 
Note that, discarding the constant number of clusters $K$, and assuming $\mathbf{W}$ is a positive semi-definite (p.s.d) affinity matrix, one can show that the objective is concave. Therefore, the first-order approximation gives the following linear upper bound for concave ${\cal F}(\Sbold)$ at current iteration $i$:
\begin{eqnarray}
{\cal H}_i(\Sbold)&=& \sum_{k} \nabla {\cal F}(S_k)^tS_k \nonumber \\
                &=& \sum_{k}\left(\mathbf{d}\frac{(S_k^i)^t\mathbf{W}S_k^i}{(\mathbf{d}^t S_k^i)^2} -\frac{2\mathbf{W}S_k^i}{\mathbf{d}^t S_k^i}\right)^t S_k \nonumber\\
                &=& \sum_{p =1}^{N} \sbold_p^t \abold_p^i \nonumber
\end{eqnarray}
with $a_{p,k}^i = d_p \frac{(S_k^i)^t\mathbf{W}S_k^i}{(\mathbf{d}^t S_k^i)^2} -\frac{2\sum_q w(\xx_p,\xx_q)s_{p,k}^i}{\mathbf{d}^t S_k^i}$, with degree vector $\mathbf{d} = [d_p]$, $d_p= \sum_q w(\xx_p,\xx_q)~\forall p$, and for all p.s.d. affinity matrices $\mathbf{W} = [w(\xx_p,\xx_q)]$.

\section{Output clusters with respect to $\lambda$. }
In Fig.\ref{fig:lambda_vs_clusters}, we plot the output clusters of Fair K-means with respect to an increased value of $\lambda$, for the synthetic data sets. When $\lambda = 0$, we get the traditional clustering results of K-means without fairness. The result clearly has biased clusters, each corresponding fully to one the demographic groups, with a balance measure equal $0$. In the \textit{Synthetic} dataset, the balance increases with increased value of parameter $\lambda$ and eventually gain the desired equal balance with a certain increased value of $\lambda$. We also observe the same trend in the \textit{Synthetic-unequal} dataset, where the output clusters are found according to prior demographic distribution $U = [0.75, 0.25]$, with almost a null fairness error starting from a certain value of $\lambda$.

\bibliography{fairness}

\begin{thebibliography}{31}
\providecommand{\natexlab}[1]{#1}
\providecommand{\url}[1]{\texttt{#1}}
\providecommand{\urlprefix}{URL }
\expandafter\ifx\csname urlstyle\endcsname\relax
  \providecommand{\doi}[1]{doi:\discretionary{}{}{}#1}\else
  \providecommand{\doi}{doi:\discretionary{}{}{}\begingroup
  \urlstyle{rm}\Url}\fi

\bibitem[{Arthur and Vassilvitskii(2007)}]{arthur2007k}
Arthur, D.; and Vassilvitskii, S. 2007.
\newblock k-means++: The advantages of careful seeding.
\newblock In \emph{ACM-SIAM symposium on Discrete algorithms}, 1027--1035.
  Society for Industrial and Applied Mathematics.

\bibitem[{Backurs et~al.(2019)Backurs, Indyk, Onak, Schieber, Vakilian, and
  Wagner}]{backurs2019scalable}
Backurs, A.; Indyk, P.; Onak, K.; Schieber, B.; Vakilian, A.; and Wagner, T.
  2019.
\newblock Scalable fair clustering.
\newblock \emph{International conference on machine learning (ICML)} 405--413.

\bibitem[{Bera et~al.(2019)Bera, Chakrabarty, Flores, and
  Negahbani}]{bera2019fair}
Bera, S.; Chakrabarty, D.; Flores, N.; and Negahbani, M. 2019.
\newblock Fair algorithms for clustering.
\newblock In \emph{Advances in Neural Information Processing Systems},
  4955--4966.

\bibitem[{Buolamwini and Gebru(2018)}]{buolamwini2018gender}
Buolamwini, J.; and Gebru, T. 2018.
\newblock Gender shades: Intersectional accuracy disparities in commercial
  gender classification.
\newblock In \emph{Conference on Fairness, Accountability and Transparency},
  77--91.

\bibitem[{Celis et~al.(2018)Celis, Keswani, Straszak, Deshpande, Kathuria, and
  Vishnoi}]{Celis2018}
Celis, L.~E.; Keswani, V.; Straszak, D.; Deshpande, A.; Kathuria, T.; and
  Vishnoi, N.~K. 2018.
\newblock Fair and Diverse DPP-Based Data Summarization.
\newblock In \emph{International Conference on Machine Learning (ICML)},
  715--724.

\bibitem[{Chierichetti et~al.(2017)Chierichetti, Kumar, Lattanzi, and
  Vassilvitskii}]{Chierichetti2017}
Chierichetti, F.; Kumar, R.; Lattanzi, S.; and Vassilvitskii, S. 2017.
\newblock Fair Clustering Through Fairlets.
\newblock In \emph{Neural Information Processing Systems (NeurIPS)},
  5036--5044.

\bibitem[{Csiszar and K{\"o}rner(2011)}]{csiszar2011information}
Csiszar, I.; and K{\"o}rner, J. 2011.
\newblock \emph{Information theory: coding theorems for discrete memoryless
  systems}.
\newblock Cambridge University Press.

\bibitem[{Donini et~al.(2018)Donini, Oneto, Ben{-}David, Shawe{-}Taylor, and
  Pontil}]{Donini2018}
Donini, M.; Oneto, L.; Ben{-}David, S.; Shawe{-}Taylor, J.; and Pontil, M.
  2018.
\newblock Empirical Risk Minimization Under Fairness Constraints.
\newblock In \emph{Neural Information Processing Systems (NeurIPS)},
  2796--2806.

\bibitem[{Dua and Graff(2017)}]{Dua:2019}
Dua, D.; and Graff, C. 2017.
\newblock {UCI} Machine Learning Repository.

\bibitem[{Hardt, Price, and Srebro(2016)}]{Hardt2016}
Hardt, M.; Price, E.; and Srebro, N. 2016.
\newblock Equality of Opportunity in Supervised Learning.
\newblock In \emph{Neural Information Processing Systems (NeurIPS)},
  3315--3323.

\bibitem[{Huang, Jiang, and Vishnoi(2019)}]{huang2019coresets}
Huang, L.; Jiang, S.; and Vishnoi, N. 2019.
\newblock Coresets for clustering with fairness constraints.
\newblock In \emph{Advances in Neural Information Processing Systems},
  7587--7598.

\bibitem[{Julia et~al.(2016)Julia, Larson, Mattu, and Kirchner}]{Julia:2016}
Julia, A.; Larson, J.; Mattu, S.; and Kirchner, L. 2016.
\newblock Propublica -- machine bias.

\bibitem[{Kleinberg et~al.(2017)Kleinberg, Lakkaraju, Leskovec, Ludwig, and
  Mullainathan}]{kleinberg2017human}
Kleinberg, J.; Lakkaraju, H.; Leskovec, J.; Ludwig, J.; and Mullainathan, S.
  2017.
\newblock Human decisions and machine predictions.
\newblock \emph{The quarterly journal of economics} 133(1): 237--293.

\bibitem[{Kleindessner et~al.(2019)Kleindessner, Samadi, Awasthi, and
  Morgenstern}]{Kleindessner2019}
Kleindessner, M.; Samadi, S.; Awasthi, P.; and Morgenstern, J. 2019.
\newblock Guarantees for Spectral Clustering with Fairness Constraints.
\newblock In \emph{International Conference of Machine Learning (ICML)},
  3458--3467.

\bibitem[{Moro, Cortez, and Rita(2014)}]{moro2014data}
Moro, S.; Cortez, P.; and Rita, P. 2014.
\newblock A data-driven approach to predict the success of bank telemarketing.
\newblock \emph{Decision Support Systems} 62: 22--31.

\bibitem[{Narasimhan and Bilmes(2005)}]{Narasimhan2005}
Narasimhan, M.; and Bilmes, J. 2005.
\newblock A Submodular-supermodular Procedure with Applications to
  Discriminative Structure Learning.
\newblock In \emph{Conference on Uncertainty in Artificial Intelligence (UAI)},
  404--412.
\newblock ISBN 0-9749039-1-4.

\bibitem[{Nocedal and Wright(2006)}]{nocedal2006numerical}
Nocedal, J.; and Wright, S. 2006.
\newblock \emph{Numerical optimization}.
\newblock Springer.

\bibitem[{R{\"o}sner and Schmidt(2018)}]{Rosner2018PrivacyPC}
R{\"o}sner, C.; and Schmidt, M. 2018.
\newblock Privacy preserving clustering with constraints.
\newblock In \emph{ICALP}.

\bibitem[{Samadi et~al.(2018)Samadi, Tantipongpipat, Morgenstern, Singh, and
  Vempala}]{Samadi2018}
Samadi, S.; Tantipongpipat, U.~T.; Morgenstern, J.~H.; Singh, M.; and Vempala,
  S. 2018.
\newblock The Price of Fair {PCA:} One Extra dimension.
\newblock In \emph{Neural Information Processing Systems (NeurIPS)},
  10999--11010.

\bibitem[{Schmidt, Schwiegelshohn, and Sohler(2018)}]{Schmidt2018}
Schmidt, M.; Schwiegelshohn, C.; and Sohler, C. 2018.
\newblock Fair Coresets and Streaming Algorithms for Fair k-Means Clustering.
\newblock \emph{arXiv 1304.6478} abs/1812.10854.

\bibitem[{Shaham et~al.(2018)Shaham, Stanton, Li, Basri, Nadler, and
  Kluger}]{shaham2018spectralnet}
Shaham, U.; Stanton, K.; Li, H.; Basri, R.; Nadler, B.; and Kluger, Y. 2018.
\newblock SpectralNet: Spectral Clustering using Deep Neural Networks.
\newblock In \emph{International Conference on Learning Representations
  (ICLR)}.

\bibitem[{Tang et~al.(2019)Tang, Marin, Ayed, and Boykov}]{Tang2019KernelCK}
Tang, M.; Marin, D.; Ayed, I.~B.; and Boykov, Y. 2019.
\newblock Kernel Cuts: Kernel and Spectral Clustering Meet Regularization.
\newblock \emph{International Journal of Computer Vision} 127: 477--511.

\bibitem[{Tian et~al.(2014)Tian, Gao, Cui, Chen, and Liu}]{Tian-AAAI}
Tian, F.; Gao, B.; Cui, Q.; Chen, E.; and Liu, T.-Y. 2014.
\newblock Learning deep representations for graph clustering.
\newblock In \emph{AAAI Conference on Artificial Intelligence}, 1293--1299.

\bibitem[{Vaida(2005)}]{vaida2005parameter}
Vaida, F. 2005.
\newblock Parameter convergence for EM and MM algorithms.
\newblock \emph{Statistica Sinica} 15: 831--840.

\bibitem[{Vladymyrov and Carreira-Perpi{\~n}{\'a}n(2016)}]{Vladymyrov-2016}
Vladymyrov, M.; and Carreira-Perpi{\~n}{\'a}n, M. 2016.
\newblock The Variational Nystrom method for large-scale spectral problems.
\newblock In \emph{International Conference on Machine Learning (ICML)},
  211--220.

\bibitem[{Von~Luxburg(2007)}]{VonLuxburg2007}
Von~Luxburg, U. 2007.
\newblock A tutorial on spectral clustering.
\newblock \emph{Statistics and computing} 17(4): 395--416.

\bibitem[{Yuan et~al.(2017)Yuan, Yin, Bai, Feng, and Tai}]{Yuan2017}
Yuan, J.; Yin, K.; Bai, Y.; Feng, X.; and Tai, X. 2017.
\newblock Bregman-Proximal Augmented Lagrangian Approach to Multiphase Image
  Segmentation.
\newblock In \emph{Scale Space and Variational Methods in Computer Vision
  (SSVM)}, 524--534.

\bibitem[{Yuille and Rangarajan(2001)}]{Yuille2001}
Yuille, A.~L.; and Rangarajan, A. 2001.
\newblock The Concave-Convex Procedure {(CCCP)}.
\newblock In \emph{Neural Information Processing Systems ({NIPS})}, 1033--1040.

\bibitem[{Zafar et~al.(2017)Zafar, Valera, Gomez{-}Rodriguez, and
  Gummadi}]{Zafar2017}
Zafar, M.~B.; Valera, I.; Gomez{-}Rodriguez, M.; and Gummadi, K.~P. 2017.
\newblock Fairness Constraints: Mechanisms for Fair Classification.
\newblock In \emph{International Conference on Artificial Intelligence and
  Statistics (AISTATS)}, 962--970.

\bibitem[{Zhang, Kwok, and Yeung(2007)}]{Zhang2007}
Zhang, Z.; Kwok, J.~T.; and Yeung, D.-Y. 2007.
\newblock Surrogate maximization/minimization algorithms and extensions.
\newblock \emph{Machine Learning} 69: 1--33.

\bibitem[{Ziko, Granger, and Ayed(2018)}]{ziko2018scalable}
Ziko, I.; Granger, E.; and Ayed, I.~B. 2018.
\newblock Scalable Laplacian K-modes.
\newblock In \emph{Advances in Neural Information Processing Systems},
  10041--10051.

\end{thebibliography}
\end{document}